\documentclass[lettersize, journal]{IEEEtran}  % Comment this line out if you need a4paper

\IEEEoverridecommandlockouts                              % This command is only needed if 
\usepackage{amssymb}
\usepackage{tabularray}
\usepackage{amsmath}
\usepackage{mathtools}
\usepackage{algorithm}
\usepackage{algpseudocode}
\usepackage{tikz}
\usepackage{layouts}
\usepackage{todonotes}
\usetikzlibrary{shapes.geometric, arrows}
\usepackage{xcolor}
\usepackage{url}
\usepackage{lipsum}
\usepackage{cite}
\usepackage{hyperref}

\usepackage{tensor}
\usepackage{stackengine}
\usepackage{hyperref,etoolbox}
\usepackage{fnpos}
\usepackage{fancyhdr}
\usepackage{pifont}
\usepackage{setspace}
\usepackage{multicol}
\usepackage{multirow}
\usepackage{graphicx}
\usepackage{makecell}
\usepackage{float}
\usepackage{stfloats}
\usepackage{booktabs}
\usepackage[font=footnotesize, labelfont=bf]{caption}
\usepackage{subcaption}
\usepackage{diagbox}
\usepackage{makecell}

\makeatletter
\patchcmd{\ALG@step}{\addtocounter{ALG@line}{1}}{\refstepcounter{ALG@line}}{}{}
\newcommand{\ALG@lineautorefname}{Line}
\makeatother

\newcommand{\pnorm}[2]{\left\lVert#1\right\rVert_{#2}}

\newcommand{\Real}{\mathbb R}

\newcommand{\task}{\mathcal{T}}
\newcommand{\q}{\mathbf{q}}
\renewcommand{\v}{\mathbf{v}}
\newcommand{\x}{\mathbf{x}}

\renewcommand{\u}{\mathbf{u}}
\newcommand{\X}{\mathcal{X}}
\newcommand{\U}{\mathcal{U}}

\newcommand{\A}{\mathbf{A}}
\newcommand{\B}{\mathbf{B}}

\newcommand{\pt}{\mathbf{P}}     % point
\newcommand{\f}{\mathbf{f}}     % function
\newcommand{\pose}{\mathbf{X}}     % 3d Pose
\newcommand{\rot}{\mathbf{R}}     % 3d Pose

\newcommand{\SE}[1]{\text{SE}(#1)}

\newcommand{\SO}[1]{\text{SO}(#1)}
\newcommand{\so}[1]{\mathfrak{so}(#1)}
\newcommand{\Log}{\operatorname{Log}}

\newcommand{\defeq}{\vcentcolon=}
\newcommand{\Q}{\mathbf{Q}}

\DeclareMathAlphabet\mathbfcal{OMS}{cmsy}{b}{n}

\DeclareMathOperator*{\argmin}{arg\,min}
\DeclareMathOperator*{\lexmin}{lex\,min}

\definecolor{ashgrey}{rgb}{0.7, 0.75, 0.71}

\def\Vec#1{\!\!\hbox{$#1$\kern-0.38em\lower0.85em\hbox{$\vec{}\,$}}\,}%
\newcommand{\bbm}{\begin{bmatrix}}
	\newcommand{\ebm}{\end{bmatrix}}
\DeclareMathAlphabet{\mbf}{OT1}{ptm}{b}{n}

\usepackage{xcolor}
\newcommand{\revision}[1]{{\color{black} #1}}

\newtheorem{preexample}{\sf\bfseries Example}

\newtheorem{prethm}{\sf\bfseries Theorem}

\newtheorem{prelem}{\sf\bfseries Lemma}

\newtheorem{preprop}{\sf\bfseries Proposition}

\makeatletter
\newenvironment{practitionersnote}
  {\abstract}
  {\endabstract}
\makeatother

\author{
    Xintong Du$^{*1}$, Jingxing Qian$^{*1}$, Siqi Zhou$^{2,3}$, Angela P. Schoellig$^2$
    \thanks{$^*$equal contribution.}
    \thanks{$^{1}$University of Toronto Institute for Aerospace Studies (UTIAS) and the Vector Institute for Artificial Intelligence, Canada.}%
    \thanks{$^2$Learning Systems and Robotics lab at the Technical University of Munich and the Munich Institute for Robotics and Machine Intelligence (MIRMI), Germany.}
    \thanks{$^3$School of Computing Science, Faculty of Applied Sciences, Simon Fraser University, Burnaby, BC, Canada.}
    \thanks{Emails: \url{xintong.du@robotics.utias.utoronto.ca}, \url{jingxing.qian@robotics.utias.utoronto.ca}, \url{angela.schoellig@tum.de}, \url{siqi@sfu.ca}.}
}

\title{\LARGE \bf
Perceptive Hierarchical-Task MPC for Sequential Mobile Manipulation in Unstructured Semi-Static Environments
}

\begin{document}

\maketitle

\setcounter{topnumber}{1}
\renewcommand{\topfraction}{0.99}
\renewcommand{\textfraction}{0.01}

%%%%%%%%%%%%%%%%%%%%%%%%%%%%%%%%%%%%%%%%%%%%%%%%%%%%%%%%%%%%%%%%%%%%%%%%%%%%%%%%
\begin{figure*}[b]
\centering
\includegraphics[width=\textwidth]{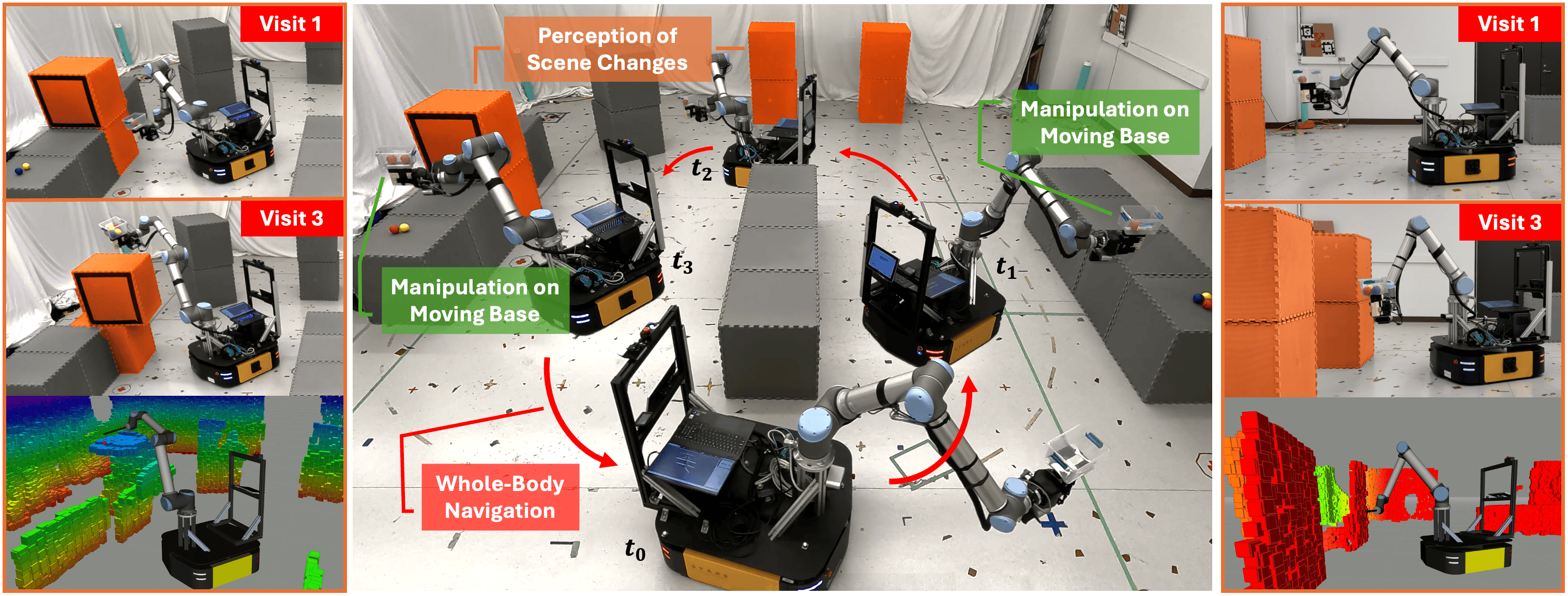}
  \caption{Demonstration of our proposed closed-loop perceptive HTMPC control system performing sequential mobile manipulation tasks in semi-static environments. The left and right panels highlight the robot’s response to changed boxes between visits, along with a snapshot of its updated 3D map, illustrating our system's capability to handle semi-static changes in the scene via online perception and motion planning. A video of the demo is available at \href{http://tiny.cc/peception-htmpc}{http://tiny.cc/peception-htmpc}.} 
  \label{fig:main}
\end{figure*}

\setlength{\textfloatsep}{10pt} % Adjusts space below the figure

\begin{abstract}

As compared to typical mobile manipulation tasks, sequential mobile manipulation poses a unique challenge---as the robot operates over extended periods, successful task completion is not solely dependent on consistent motion generation but also on the robot's awareness and adaptivity to changes in the operating environment. While existing motion planners can generate whole-body trajectories to complete sequential tasks, they typically assume that the environment remains static and rely on precomputed maps. This assumption often breaks down during long-term operations, where semi-static changes such as object removal, introduction, or shifts are common. In this work, we propose a novel perceptive hierarchical-task model predictive control (HTMPC) framework for efficient sequential mobile manipulation in unstructured, changing environments. To tackle the challenge, we leverage a Bayesian inference framework to explicitly model object-level changes and thereby maintain a temporally accurate representation of the 3D environment; this up-to-date representation is embedded in a lexicographic optimization framework to enable efficient execution of sequential tasks. We validate our perceptive HTMPC approach through both simulated and real-robot experiments. In contrast to baseline methods, our approach systematically accounts for moved and phantom obstacles, successfully completing sequential tasks with higher efficiency and reactivity, without relying on prior maps or external infrastructure.

\end{abstract}

\begin{practitionersnote}
Many real-world mobile manipulation applications, such as warehousing or service robotics, are sequential in nature---robots are required to execute a predefined sequence of tasks to achieve a high-level objective. Previous methods can efficiently coordinate whole-body motions for the redundant system, but many of them assume static environments. In operational settings, environments often undergo semi-static changes where objects are relocated, removed, or introduced over time, rendering any preplanned trajectories suboptimal or even unsafe at execution time. To enable reliable long-term autonomy, robots must continuously adapt their motion plans based on onboard perception rather than fixed environment representations. This work presents a novel perceptive control framework that bases whole-body control on an up-to-date map to support reactive behaviours for long-term operation while exploiting redundancy for efficient task execution. Our approach relies solely on onboard sensing and compute, enabling fully autonomous execution in previously unknown environments. We demonstrate our approach in representative sequential pick-and-place tasks. The modularity of our framework can be further leveraged in combination with, for instance, language-based task planners to solve more complex tasks, or with stack-of-task methods to enforce priorities that are not temporally defined. 
%  does not rely on any external localization infrastructure,
%The framework is modular and intended for practical deployment. 
% ORIGINAL
% Many practical mobile manipulation tasks are inherently sequential, requiring the robot to execute a predefined sequence of subtasks to achieve a high-level objective. Previous methods can efficiently coordinate whole-body motions for the redundant system but many of them assume static environments. The assumption often breaks down during long-term operations, where semi-static changes such as object displacement, temporary clutter, or furniture reconfiguration are common. Consequently, preplanned trajectories can become invalid or unsafe at execution time. Therefore, robots must continuously adapt their motion plans based on onboard perception rather than fixed environment representations. This work presents a novel perceptive control framework that bases whole-body control on an up-to-date map to support reactive behaviours for long-term operation while still exploiting redundancy for efficient task execution. Moreover, the system only uses onboard sensing and does not rely on any external localization infrastructure, enabling fully autonomous execution in previously unknown environments. The framework is modular and intended for practical deployment. While we demonstrate it in sequential pick-and-place tasks, the framework can be used in combination with, for instance, a language-based task planner to solve more complex tasks, or with a stack-of-task framework to enforce priorities that are not temporally defined. 
\end{practitionersnote}

%  Mobile manipulators are often required to execute long sequences of manipulation tasks at distant locations in environments that change over time. While hierarchical control methods can efficiently coordinate whole-body motions for redundant robots, their performance in practice depends critically on the availability of accurate and up-to-date environment maps. However, many existing approaches rely on offline maps or assume perfect knowledge of the environment.
% This work presents a perception-based control framework that integrates hierarchical control with object-aware mapping to support long-term operation in environments with semi-static changes, such as object displacement or temporary clutter. By maintaining object-level environment representations and updating them online from onboard perception, the system can react to environmental changes without relying on offline maps or external localization infrastructure while still exploiting redundancy for efficient task execution.
% The framework is modular and intended for practical deployment. While we demonstrate our proposed perceptive HTMPC in sequential pick-and-place tasks, the framework can be used in combination with, for instance, a language-based task planner to solve more complex tasks, or with an active perception approach to enforce priorities that are not temporally defined. 

%\setlength{\textfloatsep}{10pt}% Remove \textfloatsep

\section{Introduction}\label{sec:intro}
\revision{Mobile manipulation has gained increasing attention due to its potential to perform complex tasks that require both navigation and physical object interaction. Many practical mobile manipulation problems are sequential in nature: given high-level instructions, a robot is required to execute an ordered sequence of interleaved navigation and manipulation actions over an extended period of time~\cite{ahn_as_2022,thakar_manipulator_2022}. For instance, in a warehouse, robot may be repetitively tasked with  navigating to different storage racks to retrieve items and transport them to a packing station; similarly, in a service setting, a robot might sequentially visit different rooms to tidy up objects or deliver supplies.
%Typical examples include material transport, pick-and-place logistics, and inspection tasks, where the robot must repeatedly navigate between sites of interests and perform manipulation actions.

% A key challenge in sequential mobile manipulation lies in generating coordinated whole-body motions that are kinematically consistent during task transitions. To address this, 
There has been significant progress in sequential mobile manipulation, with a particular emphasis on generating coordinated whole-body motions that remain kinematically consistent during task transitions. Existing approaches can be broadly categorized into two main streams: heuristic-based methods and hierarchical optimization frameworks. Heuristic approaches typically rely on predefined coordination rules or reactive schemes to manage the redundancy between the base and the arm~\cite{thakar_manipulator_2022, burgess-limerick_reactivebasecontrol_2023, burgess-limerick_enabling_2023, burgess-limerick_architecture_2023}. While these methods have shown success in specific scenarios, hierarchical task control frameworks, such as hierarchical-task model predictive control (HTMPC), offer a more principled alternative~\cite{du_hierarchicaltaskmodel_2024a} by explicitly exploiting kinematic redundancy to improve execution efficiency and consistency.

Despite empirical success, existing works in either stream often make an assumption that limits their applicability in practical settings. In particular, most prior works assume static environments and thereby leverage an offline-generated prior map for downstream decision-making~\cite{thakar_manipulator_2022, burgess-limerick_reactivebasecontrol_2023, burgess-limerick_enabling_2023, burgess-limerick_architecture_2023,wu_realtimewholebodymotion_2024, zimmermann_go_2021}. However, in many real-world applications, the operating environment is subjected to semi-static changes, where objects may be shifted, added, or removed during task execution. This discrepancy is especially pronounced in long-horizon sequential tasks, where changes can occur between subtasks without immediate observation, causing the prior map to drift from reality over time. 
%sequential tasks which inherently span extended operations. 
In a warehouse context, goods pallets may be relocated onto the robot's path as it retrieves a package and returns to the packing location, while in a domestic setting, furniture can be moved around without immediate observations. In such scenarios, pre-generated plans based on the static environment assumption or offline maps may become suboptimal or even unsafe as the environment changes. To ensure reactive and efficient operation, the robot must continuously perceive its surroundings, track changes in unstructured scenes, and adapt to the changing environment online. An example of this process is presented in \autoref{fig:main}.

In this work, we propose a novel perceptive HTMPC control framework for sequential mobile manipulation in unstructured semi-static environments (\autoref{fig:system_overview}). To enable long-term autonomous operation in changing environments, we close the loop between online environment understanding and whole-body control in a coherent framework. To maintain a temporally consistent map, we explicitly model the  object-level stationarity score and leverage a Bayesian inference approach to account for changes that are possibly not directly observed; based on the updated map, we formulate a lexicographic optimization problem that enables reliable execution of sequential mobile manipulation tasks, in unstructured environments, even as the environments evolves.
%To enable long-term autonomous operation in changing environments, we close the loop between online environment understanding and whole-body control by modeling object-level semi-static changes within a Bayesian inference framework and incorporating the resulting 3D environment map into a lexicographic optimization control framework. This formulation enables efficient and reliable execution of sequential mobile manipulation tasks while performing obstacle avoidance despite environmental changes.
%  To enable long-term autonomy, our approach explicitly models object-level semi-static changes within a Bayesian inference framework and thereby maintains a consistent 3D environment map. Moreover, our approach leverages lexicographic
% optimization to exploit the kinematic redundancy of the
% robot system for efficient sequential mobile manipulation.
We demonstrate our approach in both simulation and real-robot experiments, showing that our framework outperforms typical mobile manipulation baselines in both efficiency and reactivity, without relying on a prior map or external localization infrastructure. To the best of our knowledge, this is the first work that systematically closes the perception-action loop for sequential mobile manipulation tasks in unstructured, semi-static environments, which paves the way for reliable long-term autonomy. % for efficient and reliable  

}

\section{Related Work}\label{sec:relatedwork}
%\subsection{Mobile Manipulation in Unstructured Environments}

\subsection{Sequential Mobile Manipulation}
Efficient sequential mobile manipulation requires optimal whole-body motion plans for a sequence of tasks as opposed to the simple concatenation of individual plans for each task. The key challenge is to tackle the inherent kinematic inconsistency in the desired task sequence when generating a feasible whole-body plan.   
Typical weighted-sum (WS) methods are used to handle multiple control objectives, but can produce compromised solutions that may violate task ordering. 

Some works tackle this challenge from the perspective of task-space planning~\cite{thakar_manipulator_2022, burgess-limerick_architecture_2023, burgess-limerick_enabling_2023}. In these works, heuristics are carefully hand-designed to reduce conflicts between consecutive cartesian motion plans. Others adopt more principled control formulations grounded in robot dynamics or kinematics~\cite{zimmermann_go_2021, du_hierarchicaltaskmodel_2024a}. In particular, the HTMPC framework formulates the sequential mobile manipulation task problem as a lexicographic optimization problem, providing a principled approach to enforce task sequence while leveraging the robot's redundancy~\cite{du_hierarchicaltaskmodel_2024a}.

Similarly, the HTMPC framework assumes perfect knowledge about the environment.  
Tasks are specified using static, predefined geometric locations, rather than being informed by real-time perceptions. As a result, these methods lack the ability to adapt to semi-static environments where continuous perception and change detection are essential for safe and effective task execution. In contrast to previous works, our perceptive HTMPC framework systematically accounts for semi-static changes online without relying on any prior knowledge or external infrastructures.

\subsection{Object-Aware Mapping in Changing Environments}\label{sec:lit-scene-changes}

% Most existing mapping systems are designed under the static world assumption and primarily focus on achieving high geometric accuracy. 
Earlier mapping systems primarily focus on providing an accurate geometric representation of the environments. 
These methods typically rely on sparse feature-based representations~\cite{ORBSLAM3_TRO} or dense grid-based reconstructions such as occupancy grids~\cite{octomap} and Euclidean distance fields (EDFs)~\cite{voxblox}. 
To improve scene understanding, more recent works have incorporated semantic and object-level information into geometric maps. Systems such as Voxblox++~\cite{grinvald2019volumetric}, Kimera~\cite{Rosinol20icra-Kimera}, Hydra~\cite{hughes2022hydra}, and ConceptFusion~\cite{conceptfusion} extend geometric maps with object and semantic labels, enabling semantic-level reasoning~\cite{Gu2023ConceptGraphsO3, opennav}. While these approaches provide rich geometric and semantic information, most of these approaches assume a stationary world depsite that the environment is subject to changes especially for long-term operations. As a result, these methods become vulnerable due to map corruption and degraded localization, putting the robot's safety at risks.

Several strategies have been proposed to deal with dynamic scenes. One approach identifies potential dynamics and excludes them during mapping~\cite{Schmid2023DynabloxRD}, which can lead to information loss in highly dynamic scenes. Another approach jointly tracks the robot and dynamic objects~\cite{Xu2019MIDFusionOO}, but typically relies on short-term motion consistency and is less effective for semi-static changes. More recent work explicitly targets semi-static environments. Panoptic Multi-TSDF~\cite{panoptictsdf} estimates object consistency based on the overlap between measurements and mapped objects. NeuSE~\cite{Fu2023NeuSENS} builds local scene graphs to predict object-level changes. Bayesian methods such as POCD~\cite{QianChatrathPOCD} fuses semantic priors with geometric consistency to estimate object consistency over time. However, previous mapping works rarely evaluate in closed-loop with the controller on real robots. In this work, we leverage the EDF-based volumetric mapping and object-aware change detection approach in~\cite{QianChatrathPOCD} to inform reactive control for sequential mobile manipulation tasks in cluttered, semi-static environments.

\subsection{Perceptive Control for Collision Avoidance}\label{sec:lit-mm-in-unstructured-env}

Collision avoidance is a fundamental requirement in robotic control, as it underpins safe operation in real-world environments. In inverse differential kinematics–based hierarchical task control, collision avoidance is commonly treated as a task within the hierarchy~\cite{kanoun_kinematic_2011, escande_hierarchical_2014}. In optimization-based control approaches, such as model predictive control (MPC), collision avoidance is typically enforced through explicit constraints. Recently, safety certificates, including Barrier Lyapunov functions~\cite{Tee2009BarrierLF} and CBFs~\cite{ames2019control}, have gained substantial traction as tools for enforcing safety in robotic systems. Some works instantiate CBFs as purely reactive safety filters~\cite{amesControlBarrierFunction2014}, e.g. QPs that minimally modify a nominal command while others as constraints within a MPC framework~\cite{zeng_safetycriticalmodelpredictive_2021}. Regardless of the specific formulation, these early collision-avoidance methods typically assume that collision avoidance specifications can be derived a priori from an offline map or pre-specified geometry.

More recently, perceptive safe control has emerged, in which collision avoidance are specified online from onboard sensing, enabling reactive behaviors in previously unknown and/or changing environments~\cite{pankert_perceptive_2020, burgess-limerick_reactivebasecontrol_2023, spahn_coupledmobilemanipulation_2021,zhouControlBarrierAidedTeleoperationVisualInertial2024a,chenControlBarrierFunction2025a, QianMPCCBF,liangPointCloudBasedControl2025, tanGaussianProcessesEllipsoidal2025,daiSailingPointClouds2024, desaPointCloudBasedControl2024,abdiSafeControlUsing2023}. These works can be broadly categorized by their choice of environment representation. Mapless methods derive safety constraints directly from instantaneous sensor observations~\cite{burgess-limerick_reactivebasecontrol_2023,spahn_coupledmobilemanipulation_2021, daiSailingPointClouds2024, desaPointCloudBasedControl2024,abdiSafeControlUsing2023}. While computationally efficient, such approaches inherently react to obstacles within the current sensor Field of View (FOV); this limitation can be problematic for robots with restricted sensor coverage or when occlusions are frequent. In contrast, map-based approaches explicitly construct a consistent geometric representation of the environment~\cite{pankert_perceptive_2020,zhouControlBarrierAidedTeleoperationVisualInertial2024a,chenControlBarrierFunction2025a, QianMPCCBF,liangPointCloudBasedControl2025, tanGaussianProcessesEllipsoidal2025}, which are naturally suited to longer-horizon decision making, including sequential task execution.

In this work, we formulate collision avoidance as CBF-based safety constraints within the HTMPC framework and derive these constraints online from a volumetric map. Although related map-driven MPC–CBF approaches exist~\cite{QianMPCCBF,liangPointCloudBasedControl2025,tanGaussianProcessesEllipsoidal2025}, they are predominantly demonstrated on low-DoF platforms such as quadrotors or mobile robots. Extending to mobile manipulators poses challenges in both CBF formulation and the computational efficiency due to their higher-dimensional dynamics and more complex kinematic structures.  Moreover, existing perceptive safe-control evaluations rarely consider sequential mobile manipulation under semi-static environmental changes. We address these gaps by demonstrating the perceptive HTMPC with map-driven CBF safety constraints on mobile manipulators executing sequential tasks while adapting online to changing environments.

\section{Problem Formulation}
We study the sequential mobile manipulation problem where a robot must execute a prescribed ordered set of subtasks, $[\task_1, \dots, \task_L]$. Each subtask specifies a cartesian trajectory that encodes desired task-space behaviour for the mobile base or the end-effector (EE). We expect the robot to operate in semi-static environments where any rigid objects can appear, disappear and shift over time. We assume the robot is equipped with onboard sensors and has access to RGB-D images, wheel odometry and arm joint states for online map construction and robot localization. Given the subtask set and the onboard measurements, the key challenge is to ensure subtask completion in the given order while maintaining safety in the presence of semi-static changes in the environment.

We denote the robot's state as $\x = [\q^T, \v^T]^T$ and accelerations as $\u \in \Real^m$ with generalized coordinates $\q \in \Real^n$, the generalized velocity $\v \in \Real^m$. They all have two components, the base and the arm, denoted using the subscripts $b$ and $a$, respectively. 
% The generalized coordinate of the base consists of its position and heading $\q_b=[x,y,\theta]$, and its velocity is simply $\v_b = \dot{\q}_b$.  
We denote the feasible sets for the robot state and control inputs as $\X$ and~$\U$, respectively. $\SE{\cdot}$ denotes the Special Euclidean group; $\SO{\cdot}$ denotes the Special Orthogonal group.
\section{Methodology}\label{sec:method}
% \begin{figure}[t!]
% \centering
% \includegraphics[width=0.97\linewidth]{figure/system_rev_2.0.png}
%   \caption{An overview of the proposed perceptive control framework for sequential mobile manipulation tasks.} 
%   \label{fig:system_overview}
% \end{figure}

% \begin{figure*}[ht!]
% \centering
% \includegraphics[width=\textwidth, height=4.5cm, trim=0 0.25cm 0 0.25cm, clip]{figure/Arch.pdf}
%   \caption{Proposed perceptive HTMPC framework for sequential mobile manipulation tasks in unstructured semi-static environments. Autonomy modules are represented as blocks and key components are highlighted. Arrows indicate the direction of information flow. The Perceptive HTMPC closes the perception–control loop by using the robot state and map produced by the perception and mapping module, enabling safe and reactive behavior in semi-static environments.} \label{fig:system_overview}
% \end{figure*}

\begin{figure}[t!]
\centering
\includegraphics[width=\linewidth, trim=0 0.75cm 7.8cm 0.25cm, clip]{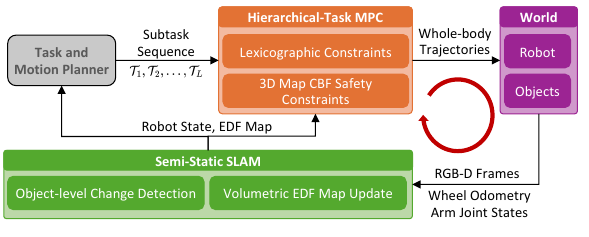}
  \caption{Proposed perceptive HTMPC framework for sequential mobile manipulation tasks in unstructured semi-static environments. Key components are highlighted. The Perceptive HTMPC closes the perception–control loop by using the robot state and map produced by the perception and mapping module, enabling safe and reactive behavior in semi-static environments.} \label{fig:system_overview}
\end{figure}

We address the aforementioned sequential mobile manipulation problem with a preceptive HTMPC framework that closes the loop between perception and whole-body control. An block diagram of our proposed approach  is shown in \autoref{fig:system_overview}. In our framework, the semi-static Simultaneous Mapping and Localization (SLAM) module estimates the current robot state~$\mathbf{x}$ and maintains an object-level 3D volumetric map $\mathcal{O}$ of the scene. From the volumetric map $\mathcal{O}$, we instantiate CBF safety constraints in the HTMPC controller online, yielding safety specifications that adapts immediately as the map is updated. Given the ordered task-space trajectories, the estimated robot state, and the latest map, HTMPC solves a lexicographic optimization problem to generate whole-body joint trajectories that are collision-free and consistent with the prescribed subtask sequence. The remainder of this section presents the detailed formulation of the proposed perceptive HTMPC framework.

% \subsection{System Overview}\label{sec:method-sys-overview}
% \input{sections/method/overview}

\subsection{Perception and Mapping}
\label{sec:method-perception&mapping}
% take rgbd from hand and base

% \begin{figure}[t!]
% \centering
% \includegraphics[width=0.75\linewidth]{figure/robot.png}
%   \caption{Our mobile manipulator setup consists of a Universal Robots UR10 arm mounted on a Clearpath Robotics Ridgeback omnidirectional base. Two synchronized Orbbec Femto Bolt RGB-D cameras are installed, with one mounted on the base and the other mounted on the end effector. An onboard laptop runs all the perception, planning and control algorithms. } 
%   \label{fig:robot_setup}
% \end{figure}

 Building on top of~\cite{QianChatrathPOCD,QianMPCCBF}, the perception and mapping module, estimates the robot's state, constructs a 3D object-level volumetric map online and updates it via continuous scene change detection. Compared to prior works using static offline volumetric maps~\cite{wu_realtimewholebodymotion_2024,burgess-limerick_architecture_2023},  our work assumes no prior scene knowledge and updates the map with a explicit scene change detection process, i.e. object matching and probabilistic change detection, to ensure safe long-term operation.

% state estimation with base camera -> modified orbslam3 with odometry, use odometry to compensate camera and compute latency, arm config from driver
\subsubsection{State Estimation}
We assume the arm's joint angles $\mathbf{q}_{a,k}$ and velocities $\mathbf{v}_{a, k}$ are readily available from the onboard joint encoders. To estimate the base pose $\mathbf{q}_{b,k}$ and velocity $\mathbf{v}_{b,k}$ at the $k$-th time step $t_k$, we modified the ORB-SLAM3~\cite{ORBSLAM3_TRO} method to use wheel odometry along with synchronized RGB-D frames from onboard cameras. The wheel odometry improves system robustness in featureless regions and extrapolates the estimated state to compensate for the low frame rate and latency of the cameras.

% While the base camera extrinsics is fixed, the EE state $\{\mathbf{q}_{a,k},\mathbf{v}_{a,k}\}$, along with the extrinsics of the attach hand camera, are estimated through arm forward kinematics using joint encoder measurements.

%A modified ORB-SLAM3~\cite{ORBSLAM3_TRO} estimates the base pose $\mathbf{q}_{b,k}$ and velocity $\mathbf{v}_{b,k}$ at $50Hz$ using the base camera frame and wheel odometry. The wheel odometry improves system robustness in featureless regions and extrapolates the estimated state to compensate for the low frame rate (15 FPS) and a $150ms$ latency of the cameras.

% $\{\mathbf{T}^{\textrm{basecam}}_{\textrm{baselink}}, \mathbf{T}^{\textrm{handcam}}_{\textrm{baselink},t}\}$
% use both base and ee camera for mapping, build on top of POCD, talk about map and object representation
\subsubsection{Object Mapping and Change Maintenance}
Given the estimated robot state $\x$, the RGB-D frames are unprojected into the global frame, segmented into pointcloud observations $\mathcal{Y}=\{\mathbf{Y}_{k,p}\}_{p=1 \cdots P}$ via geometric clustering, and merged into an object library $\mathcal{O}=\{\mathbf{O}_{i}\}_{i=1 \cdots I}$ via optimal association. Following POCD~\cite{QianChatrathPOCD}, a volumetric mapping framework for semi-static scenes, each object $\mathbf{O}_{i}$ is modeled as a submap and contains the following elements:
\begin{itemize}
    \item a 3D position and a 1D heading in the global frame,
    \item an EDF voxel grid submap, $\mathcal{M}_i$, in the global frame,
    \item a state distribution, $p(l_i,v_i)$, to model the object's current geometric change, $l_i\in\mathbb{R}$, and consistency (likelihood that the object has not changed), $v_i \in [0,1]$.
\end{itemize}

% map update
As in~\cite{QianChatrathPOCD}, the state distribution of object $\mathbf{O}_{i}$ estimated at the $k$-th time step is parameterized as the product of a Gaussian distribution for $l_{i}$ and a Beta distribution for $v_{i}$:
\begin{equation}
\label{eq:pocd_state_param}
\begin{aligned}
    p\left(l_{i}, v_{i}\right) &\defeq q\left(l_{i}, v_{i} \mid \mu_{i,k}, \sigma_{i,k}, \alpha_{i,k}, \beta_{i,k}\right)\\
        &\defeq \mathcal{N}(l_{i} \mid \mu_{i,k}, \sigma_{i,k}^2)\textrm{Beta}(v_{i} \mid \alpha_{i,k}, \beta_{i,k}).
\end{aligned}
\end{equation}
We initialize the object library as an empty set $\mathcal{O} = \varnothing$ and update it with new observations. In particular, we explicitly match new observations with the object library. Each matched object observation consists of two parts. First, we model the object geometric consistency $\Delta_{i,k}$, computed as the average difference between the object and observation EDFs. This geometric measurement can be classified as an inlier when the object has not changed, and $\Delta_{i,k}$ is normally distributed around the current estimate $l_{i}$ with standard deviation $\tau$, or as an outlier measurement where the object is moved and the change is uniformly distributed in an interval, $[-\Delta_{\textrm{max}}, \Delta_{\textrm{max}}]$. Intuitively, the geometric consistency likelihood can be parametrized as a Gaussian-Uniform mixture weighted by the object's consistency $v_{i}$:
\begin{equation}
\label{eq:pocd_depth_meas}
\begin{aligned}
&\hspace{2mm} p\left(\Delta_{i,k} \mid l_{i}, v_{i}\right) \defeq v_{i}\mathcal{N}(\Delta_{i,k} \mid l_{i}, \tau^2) \\
& \hspace{2.5cm} + (1-v_{i})\mathcal{U}(\Delta_{i,k} \mid -\Delta_{\textrm{max}}, \Delta_{\textrm{max}}).
\end{aligned}
\end{equation}
Second, we model our prior knowledge of the object classes. A binary semantic label, $s_{i,k} \in \{0,1\}$, is inferred for each observation using a classification network, where 0 denotes likely changing objects and 1 denotes likely static objects. In this work, we assume all objects are likely to change. The inferred label can be considered as a sample from a Bernoulli process controlled by the object's consistency estimate:
\begin{equation}
\label{eq:pocd_label_meas}
\begin{aligned}
&\hspace{5mm} p\left(s_{i,k} \mid v_i\right) \defeq  \textrm{Bernoulli}(s_{i,k} \mid v_{i}).
\end{aligned}
\end{equation}
Matched observations are integrated into the associated objects, and the object
state distributions, $p(l, v)$, are propagated using an approximated, closed-form Bayesian update rule derived in~\cite{QianChatrathPOCD}:

\begin{equation}
\label{eq:pocd_true_posterior}
\begin{array}{l}
    q(l_{i},v_{i} \mid \mu_{i,k+1},\sigma_{i,k+1},\alpha_{i,k+1},\beta_{i,k+1}) \simeq\\
    \quad p(\Delta_{i,k} \mid l_{i},v_{i})p(s_{i,k} \mid v_{i}) \\
    \quad\quad q(l_{i},v_{i} \mid \mu_{i,k},\sigma_{i,k},\alpha_{i,k},\beta_{i,k}).
\end{array}
\end{equation}
Unmatched observations are added to $\mathcal{O}$ as new objects. if an object's expected consistency, $\mathbb{E}[v_i]$, falls below a threshold, then the object is considered changed, and removed or relocated in the map. 

%We use both cameras for mapping to improve the map coverage, and assume the movable rigid objects in the scene only rotate around the $z$-axis. 

\vspace{-2mm}

\subsection{Perceptive HTMPC}\label{sec:method-planning&control}
 We use the HTMPC framework to generate coordinated whole-body motion plans for sequential mobile manipulation tasks. Compared to other heuristics-based motion planning works, HTMPC principally enforces the strict task sequence using the lexicographic optimization framework. In the following sections, we introduce the HTMPC framework that was first proposed in \cite{du_hierarchicaltaskmodel_2024a} but generalized in this work to include hierarchical pose tracking tasks. We complete the section with practical notes regarding control challenges due to semi-static changes. 
\subsubsection{HTMPC}
% As in \cite{du_hierarchicaltaskmodel_2024a}, we consider a mobile manipulator with a kinematic model written as
% \begin{equation}\label{eq:kinematics}
% \dot \x (t) = \A(\q(t))\x(t) + \B\u(t).
% \end{equation}
% where 
% \begin{equation}
%    \A = \begin{bmatrix}
%    \mathbf{0}_{n\times n}& \mathbf{G}(\q)\\
%    \mathbf{0}_{m\times n}&\mathbf{0}_{m\times m}
%    \end{bmatrix},\;
%     \B = \begin{bmatrix}
%    \mathbf{0}_{n\times m}\\
%    \mathbf{I}_{m\times m}
%    \end{bmatrix},
% \end{equation}
% and $\mathbf{G}(\q)$ relates the generalized coordinate and velocity by $\dot \q = \mathbf{G}(\q) \v$.
% In the following discussion, we will drop the time dependency for clarity.
% Let $\mathcal{F}_A$ be an arbitrary frame rigidly attached to a point A on the robot. Let $\pose^{A} \in \SE{3} $ be its transform with its position denoted as $\pt^A \in \Real^3$ and orientation as $ \rot^A \in \SO{3}$. Its transform as seen in the world frame can be determined from its forward kinematics
% $
% \pose^A = \f_{kin}^A(\q).
% $
% We drop the superscript A in the following formulations, which are frame-independent.
Let $\mathcal{F}_A$ be an arbitrary frame rigidly attached to a point A on the robot. Let $\pose^{A} \in \SE{3} $ be its transform with its position denoted as $\pt^A \in \Real^3$ and orientation as $ \rot^A \in \SO{3}$. Its transform as seen in the world frame can be determined from its forward kinematics
$
\pose^A = \f_{kin}^A(\q).
$
We drop the superscript A in the following formulations, which are frame-independent.

For a pose trajectory tracking task $\task$ for an arbitrary frame on the robot, the robot needs to follow a desired reference signal $\pose^d(t): [0, T] \rightarrow \SE{3}$ with the target frame on its body where $t$ is the control time and the superscript $d$ stands for the desired reference signal. The distance between $\pose$ and $\pose^d$ is denoted as $dist(\pose, \pose^d) \defeq \pnorm{\mathbf{e}}{}$ with $\pnorm{\cdot}{\cdot}$ denoting $l^2$-norm. The error vector $\mathbf{e}$ is given as 
\begin{equation}\label{eq:pose_tracking_task_distance}
    \mathbf{e} = \begin{bmatrix}
                    \mathbf{e}_{pos} \\
                    \mathbf{e}_{orn}
                 \end{bmatrix}
                 = \begin{bmatrix}
                    \pt^d - \pt \\
                    \Log(\rot^{-1} \rot^d)
                 \end{bmatrix}\\
\end{equation}
with both position error $\mathbf{e}_{pos}$ and orientation error $\mathbf{e}_{orn}$. The orientation error is determined using matrix logarithm $\Log: \SO{3} \rightarrow \so{3}$. For consistency, we also formulate the robot's base frame in the 3D space, but it only moves in the world's $x$-$y$ plane and rotates only around the $z$-axis.

HTMPC formulates the sequential tracking problem as a lexicographic optimization problem and solves it in the standard receding horizon fashion. At the $k$-th control time step $t_k$, the optimization problem, defined over a fixed time interval $[t_k, t_k+T]$, is given as
\begin{subequations}\label{eq:HT-MPC-continuous-time}
\begin{align}
      \lexmin_{\x, \u} \; &[\mathcal{J}_1, \mathcal{J}_2. \cdots, \mathcal{J}_L] \label{eq:HT-MPC-cont-cost-vec}\\
 \text{s.t.} \; & \dot\x = \A(\q) \x + \B\u, & \label{eq:HT-MPC-cont-mdl}\\
 & \mathbf{g}_{\mathit{safety}}(\x, \u) \leqslant \mathbf{0}, \label{eq:HT-MPC-cont-safetycst}\\
 & \x \in \mathcal{X}, \u \in \mathcal{U},\label{eq:HT-MPC-cont-xucst}\\
 & \x(t_k) = \x_k\label{eq:HT-MPC-cont-xocst},
\end{align}
\end{subequations}
where \eqref{eq:HT-MPC-cont-safetycst} defines the safety constraints, including self- and environmental collision avoidance. We encode environmental collision avoidance using CBF, utilizing the online-built 3D map, which is detailed in \autoref{sec:method-cbf-formulation}. The cost function for each task is the accumulated pose tracking error $\mathcal{J} = \frac{1}{2} \int_{t_k}^{t_k+T}\pnorm{\mathbf{e}}{\Q_e}^2 \,dt,$
% \begin{equation} \label{eq:HT-MPC-cont-cost-scalar}
%     \mathcal{J} = \frac{1}{2} \int_{t_k}^{t_k+T}\pnorm{\mathbf{e}}{\Q_e}^2 \,dt,
% \end{equation}
where $\Q_{e}$ is a weighting matrix of appropriate dimensions. 

Following~\cite{du_hierarchicaltaskmodel_2024a}, we solve the HTMPC problem \eqref{eq:HT-MPC-continuous-time} by solving a sequence of Single-Task MPC problems (STMPC) from the first task $\task_1$ to the last $\task_L$. 
In iteration $l$, the following optimization problem is solved to optimize $\task_l$:
\begin{subequations}\label{eq:ST-MPC-cont}
\begin{align}
      \x^{l^*}, \u^{l^*} = \argmin_{\x, \u} \; &\mathcal{J}_l + \mathcal{J}_\revision{\mathit{aux}} \\
 \text{s.t.} \, & \eqref{eq:HT-MPC-cont-mdl}, \eqref{eq:HT-MPC-cont-safetycst}, \eqref{eq:HT-MPC-cont-xucst}, \eqref{eq:HT-MPC-cont-xocst}\\
 % & {h}_i(\x, \u;{\x^i}^*, {\u^i}^* ) \leqslant 0, \forall i=1, \cdots,l-1 \label{eq:ST-MPC-cont-lexcst}\;.
 &  \mathbf{e}_i (\x, \u) \preceq_{\Real_{+}} \vert \mathbf{e}_i^* ({\x^i}^*, {\u^i}^*)\vert, \notag \\ &\hspace{1cm}\forall i=1, \cdots,l-1 \label{eq:ST-MPC-cont-lexcst}\;.
\end{align}
\end{subequations}
The additional decoupled lexicographic optimality constraints \eqref{eq:ST-MPC-cont-lexcst} limit the tracking error for each preceding task, where $\vert \; \cdot \; \vert$ denotes element-wise absolute value and $\preceq_{\Real_{+}}$ denotes element-wise inequality between two vectors.
The cost function in \eqref{eq:ST-MPC-cont} is the sum of the pose tracking error $\mathcal{J}_l$ and an auxiliary cost function for the pose tracking task
\begin{equation}\label{eq:HT-MPC-cont-aux-cost-scalar}
    \mathcal{J}_{\mathit{aux}} = \frac{1}{2} \int_{t_k}^{t_k+T}\pnorm{\dot{\mathbf{e}}}{\Q_{\dot e}}^2 + \pnorm{\x}{\Q_{x}}^2 + \pnorm{\u}{\Q_{u}}^2 \,dt,
\end{equation}
The auxiliary cost improves control performance and numerical stability, but it is not used for evaluating task hierarchy.

\subsubsection{Dealing with the Semi-Static Map Changes}
Semi-static changes in the environments impose significant challenges to the feasibility of MPC. Sudden, unexpected appearance of objects might suddenly render the MPC problem infeasible. To mitigate this problem, we softened all constraints with slack variables to allow temporary constraint violations except for the control input and lexicographic optimality constraints. The slack variables are penalized in the cost function, whose penalty parameters are set based on the exact penalty method to minimize constraint violations if feasible solutions exist. In addition to softened constraints, the proposed CBF safety constraints steer the robot away from the safety boundary if the robot is in the safe set or, if not, push the robot towards the safe set.

\subsection{Consturction of CBF Constraints from 3D Map} 
\label{sec:method-cbf-formulation}
%intro
% Given the tracked object library from perception and the HTMPC design, we would like our mobile manipulator to efficiently and safely perform navigation and manipulation tasks. However, as discussed in previous works~\cite{QianMPCCBF}, controllers relying on geometric safety constraints \eqref{eq:HT-MPC-cont-safetycst} such as the distance to obstacle surface, often violate safety constraints due to various real-world limitations, including limited sensor field of view, limited prediction horizon, presence of sensor noises, and system delays, resulting in controller infeasibility or even collision with obstacles. In this section, we discuss how the object map is transformed into a CBF to provide safety guarantees.
In perceptive HTMPC, safety constraints are constructed directly from the tracked object library $\mathcal{O}$ to enable real-time collision avoidance in semi-static environments. Due to perception limitations such as, delay, occlusion and sensor noises, additional conservativeness is required to ensure robust performance. Therefore, we propose the CBF-based formulation that encourages safe behaviours such as reducing speed near obstacles and maintaining larger traverse clearance. 

% CBF review
\subsubsection{CBF Preliminary}
The CBF certification framework~\cite{ames2019control} provides a means to ensure the safety of a system by enforcing the positive invariance of a safe set. This means that if the system starts in a safe region, it will remain in this safe region. % indefinitely, provided the system's evolution is governed by appropriate control inputs.
Let $h: \mathcal{X} \to \Real$ be a continuously differentiable function. The safe set $\mathcal{C}$ is defined as the zero-superlevel set of $h$. Specifically, the safe set $\mathcal{C}$ and its boundary $\partial \mathcal{C}$ can be written as follows:
\[
    \mathcal{C} = \{\mathbf{x} \in \mathcal{X} \mid h(\mathbf{x}) \geq 0 \}\quad\text{and} \quad \partial \mathcal{C} = \{\mathbf{x} \in \mathcal{X} \mid h(\mathbf{x}) = 0 \}.
\]
For a continuous-time system governed by the dynamics $\dot{\mathbf{x}} = f(\mathbf{x}, \mathbf{u})$, $h(\mathbf{x})$ is said to be a CBF if \textit{(i)} it satisfies $\frac{\partial h}{\partial \mathbf{x}}(\mathbf{x}) \neq \mathbf{0}$, $\forall \mathbf{x} \in \partial \mathcal{C}$, and \textit{(ii)} there exists an extended class-$\mathcal{K}_\infty$ function $\alpha:\Real\mapsto\Real$ such that  $\sup_{\mathbf{u}\in \U}\: \dot{h}(\mathbf{x},\mathbf{u}) \ge -\alpha\left( h(\mathbf{x})\right)$, $\forall \mathbf{x}\in\mathcal{X}$. 
%These conditions ensure that if the system starts in the safe set $\mathcal{C}$, there exist control inputs~$\mathbf{u}$ that can render the safe set positive invariant~\cite{ames2019control}. 
In the CBF-based MPC approach, the following constraint is typically imposed:
    $\dot{h}(\mathbf{x}, \mathbf{u}) \geq -\alpha (h(\mathbf{x}))$.
This constraint on $\mathbf{u}$ ensures the positive invariance of the safe set \cite{zeng_safetycriticalmodelpredictive_2021}.
% \begin{itemize}
% \small
%     \item \textit{Non-zero gradient on the boundary}:
%        $\frac{\partial h}{\partial \mathbf{x}}(\mathbf{x}) \neq \mathbf{0}, \quad \forall \mathbf{x} \in \partial \mathcal{C}$.

%     \item \textit{Existence of an extended class-$\mathcal{K}_\infty$ function, $\gamma:\R\mapsto\R$}:
%         $$
%             \sup_{\mathbf{u}\in \U}\: \dot{h}(\mathbf{x},\mathbf{u}) \ge -\gamma\left( h(\mathbf{x})\right),\hspace{1em} \forall \mathbf{x}\in\mathcal{X}.
%             \label{eqn:cbf_definition}
%         $$
%        %where
%       % $
%       % \dot{h}(\mathbf{x}, \mathbf{u}) = \frac{\partial h}{\partial \mathbf{x}}(\mathbf{x}) \dot{\mathbf{x}} = %\frac{\partial h}{\partial \mathbf{x}}(\mathbf{x}) f(\mathbf{x}, \mathbf{u}).
%      %  $
% \end{itemize}

%Moreover, using the simplified form $\gamma h(\mathbf{x})$ for the barrier condition allows the rate of change of $h(\mathbf{x})$ to be proportional to its current value, leading to exponential convergence toward the boundary of the safe set when necessary.

%This proportional decay condition, $\dot{h}(\mathbf{x}, \mathbf{u}) \geq -\gamma h(\mathbf{x})$, ensures that the system behaves conservatively as it approaches the boundary of the safe set but remains less restrictive when far from the boundary. This leads to smoother control actions and a more efficient control strategy, especially in applications like Model Predictive Control (MPC), where safety constraints must be maintained while optimizing system performance.

% ESDF map generation around robot
\subsubsection{EDF Local Map Construction}
We generate a local 3D EDF voxel grid $\mathcal{M}_{edf}$ centred at the robot. For each object $\mathbf{O}_i$ from the most recent object library $\mathcal{O}$, we extract a zero-level set, $\partial\mathcal{M}_{i}$, thresholded by a small value, $\theta_{\textrm{zero}}$, to approximate its surface, and take the minimum distance with respect to all objects at each voxel:
\begin{subequations}
\label{eq:esdf_map}
\begin{align}
    & \partial\mathcal{M}_{i} = \{(x,y,z) \mid |\mathcal{M}_{i}(x,y,z)| \leq \theta_{\textrm{zero}}\}, \\
    & \mathcal{M}_{edf}(x,y,z) = \min_{\substack{\mathbf{O}_i \in \mathcal{O} \\ (\bar{x},\bar{y},\bar{z})\in \partial\mathcal{M}}_{i} } \|(x,y,z)  - (\bar{x},\bar{y},\bar{z})\|_2.
\end{align}
\end{subequations}
Lastly, we truncate the map at distance $\theta_{\textit{cutoff}}$, $\mathcal{M}^{\dagger}_{\textit{edf}} = \min \{\mathcal{M}_{edf}(x,y,z), \theta_{\textrm{cutoff}}\}$.

\subsubsection{CBF Constraints}
To approximate the robot's envelope, we cover the robot arm and the base with spheres and use it as the collision model. We impose a CBF constraint for each collision sphere centered at the point $\pt^{S_j}$ with a radius $r_j$. The safe set requires the sphere to keep a desired safe clearance to the obstacles $\delta_{\textit{safe}}$. Specifically, the function that defines the safety set is given as
\begin{equation}
\label{eq:map_to_cbf}
    h_j(\mathbf{x}) = \textrm{Interp3D}(\mathcal{M}^{\dagger}_{edf})(\pt^{S_j}) - r_j - \delta_{\mathit{safe}}, \forall j = 1, 2, \dots, M
\end{equation}
where the discrete map is interpolated using a trilinear method $\textrm{Interp3D}$ to render the function $h_j$ differentiable. 

\begin{figure}[t]
    \centering
    \begin{subfigure}[b]{\linewidth}
        \includegraphics[width=\linewidth, trim=100 75 100 125, clip]{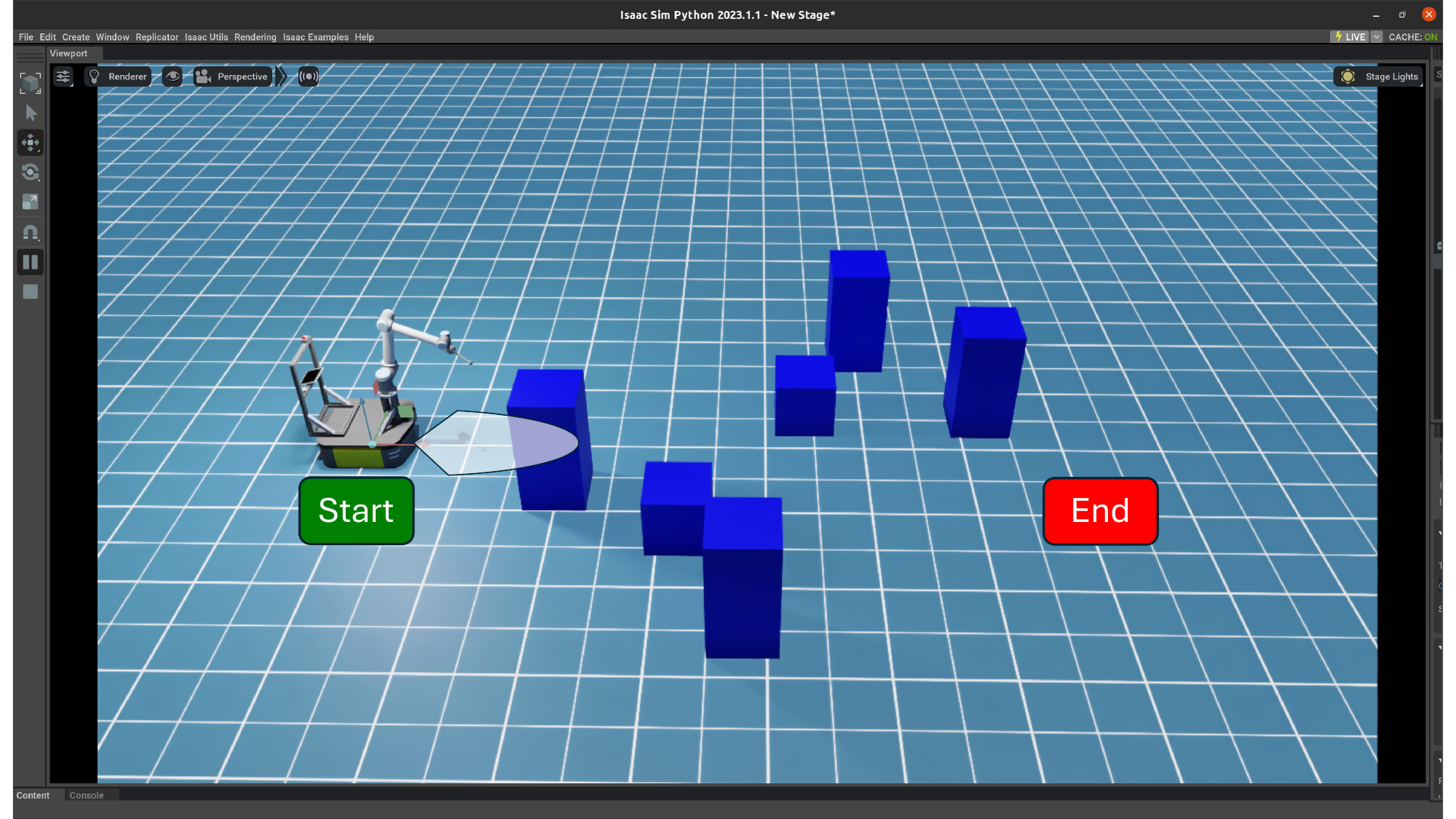}
        \caption{Phase 1: Outbound trip }
    \end{subfigure}
    \\
    \begin{subfigure}[b]{\linewidth}
        \includegraphics[width=\linewidth, trim=100 75 100 125, clip]{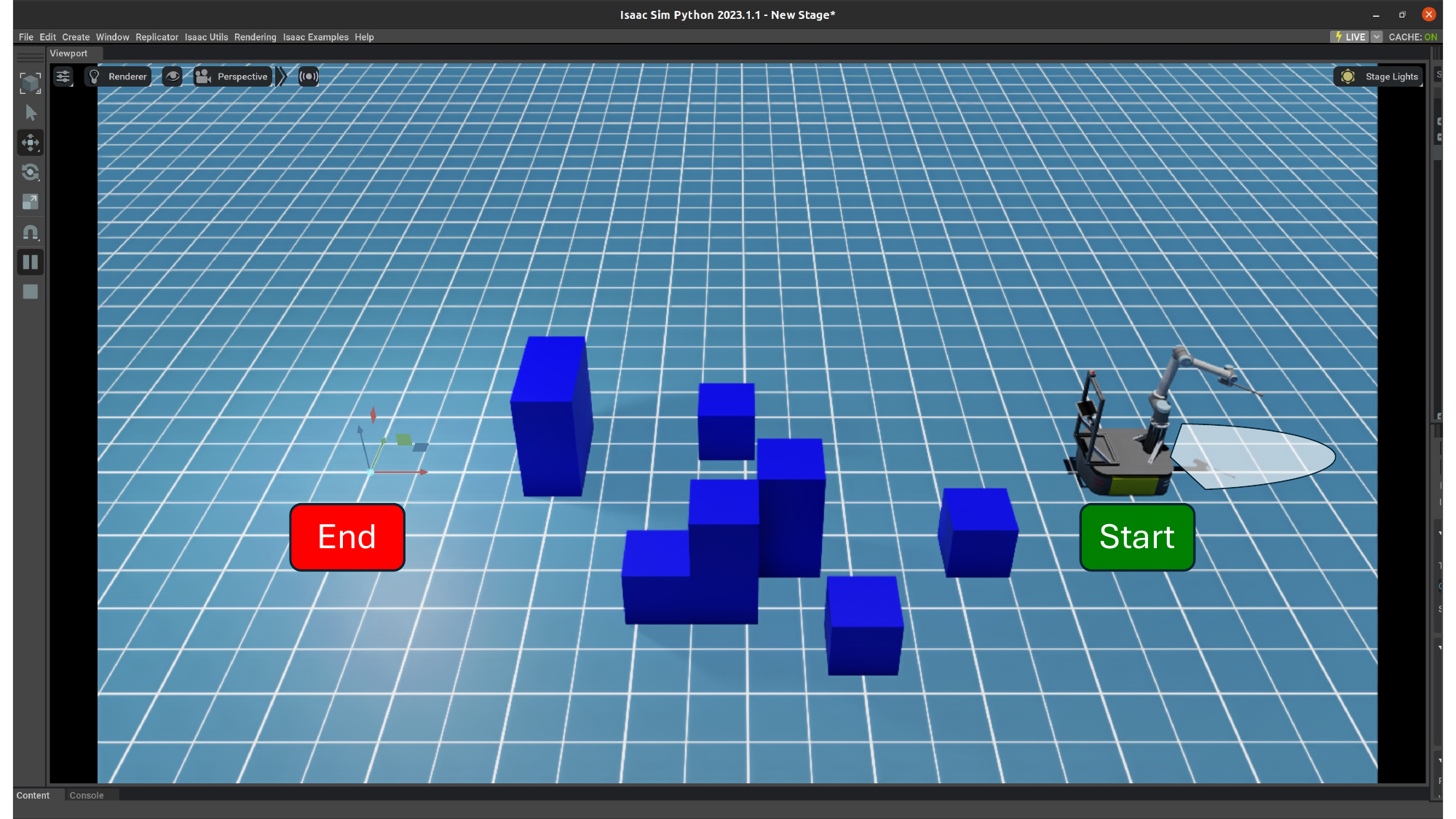}
        \caption{Phase 2: Return trip}
    \end{subfigure}
    \caption{Illustration of the whole-body navigation task in semi-static environments. The robot is tasked to make a round trip between two points. Before the robot returns, obstacles are rearranged outside the robot's FoV (white). Experiments are performed both in simulation and on the real robot.}
    \label{fig:S-WBN-Setup}
\end{figure}

\begin{figure}[t]
    \centering
    \includegraphics[width=\linewidth]{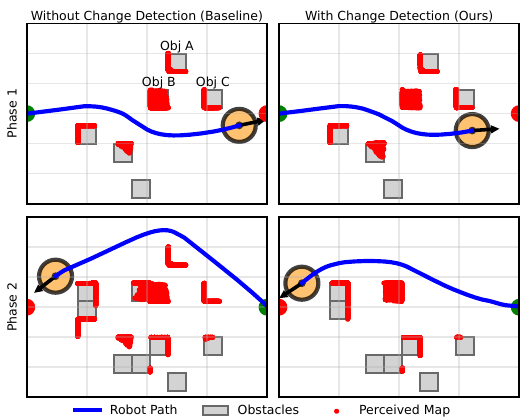}
    \caption{Comparison of the proposed mapping with change detection (right) against a baseline that assumes a static environment (left)\cite{Rosinol20icra-Kimera}. Our method produces shorter paths after scene change (phase 2) by maintaining an artifact-free map without the phantom obstacles presented in the baseline.}
    \label{fig:MS-WBN}
\end{figure}

\begin{figure}[t]
    \centering
    \includegraphics[width=\linewidth]{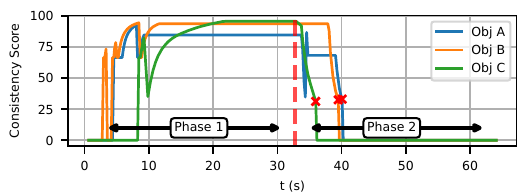}
    \caption{Time evolution of the object consistency for the three labeled objects in \autoref{fig:MS-WBN} tracked by the proposed perceptive-HTMPC framework. After scene change (red dashed line), in Phase 2, the proposed mapping system detected the relocated objects and removed them (red cross) from the map when their consistency expectation dropped below the set threshold. }
    \label{fig:MS-WBN-OBJ-CONF}
\end{figure}

Typical EDF-based safety constraints are given as $h_j(\mathbf{x})>0$. In contrast, we propose the CBF safety constraints 
$\dot{h_j}(\mathbf{x}, \mathbf{u})>-\gamma h_j(\x)$ where the extended class function is chosen to be a constant. Unlike the EDF constraints which limit the robot's position, the CBF constraints limit how fast the robot can travel towards the safety boundary or equivalently, the obstacles.
Here, $\gamma$ can be considered a decay factor in the speed limit as well as the traverse clearance. A larger $\gamma$ allows the robot to act more aggressively around obstacles, while a smaller $\gamma$ forces the robot to react more conservatively. 

% CBFs provide theoretical guarantees of safety under ideal conditions. However, even in simulation with a complete and noise-free map, a collision may still occur if the robot is not accurately modelled, if the constraints are softened to ensure feasibility, or if the optimization problem is not solved to optimality at each time step. These are practical implementation trade-offs, balancing control accuracy with real-time computational constraints. Therefore, in our evaluations, the CBF-based safety constraints occasionally leads to collisions due to these limitations. Nevertheless, as we show in the next section, the CBF-based approach outperforms the typical EDF formulation across all safety measures---most notably achieving a lower collision rate---due to its more conservative behaviours given incomplete environmental information. Furthermore, the majority of collisions of the CBF approach stem from incomplete maps rather than inadequate control performance.

\revision{
Both EDF and CBF formulations are effective in idealized settings. However, even in simulation with a complete and noise-free map, a collision may still occur if the robot is not accurately modelled, if the constraints are softened to ensure feasibility, or if the optimization problem is not solved to optimality at each time step. These are practical implementation trade-offs, balancing control accuracy with real-time computational constraints. Moreover, in reality, both methods are subjected to perception delays, occlusions and uncertainties. In such cases, CBF provides a better means to define conservative behaviours around obstacles through the extended class-$\mathcal{K}_\infty$ function\cite{QianMPCCBF}. }

% \begin{figure*}[t]
%     \centering
%     \begin{subfigure}[b]{\textwidth}
%         \includegraphics[width=\linewidth, trim=0 8 0 8, clip]{figure/benchmark/benchmark_realworld_screenshots.pdf}
%         % \caption{Caption for figure 1}
%         \label{fig:S-WBN-Imgaes}
%     \end{subfigure}
%     \\
%     \begin{subfigure}[b]{\textwidth}
%         \includegraphics[width=\linewidth, trim=0 1 0 6, clip]{figure/benchmark/benchmark_realworld_data.pdf}
%         % \caption{Caption for figure 2}
%         \label{fig:S-WBN-Rendering}
%     \end{subfigure}
%     \vspace{-3mm} % Reduce vertical spacing if needed
%     \caption{Real-world scenario of the safe whole-body navigation tasks in unstructured semi-static environments (S-WBN), highlighting real-time decision making with the proposed HTMPC-CBF approach in closed-loop with onboard perception and safe navigation despite an incomplete map. \textit{Top row}: time-series images from the real-robot experiment. \textit{Bottom row}: corresponding illustrations of the environment, tasks, the robot’s perceived map, estimated states, and motion plans.}
%     \label{fig:S-WBN-Task}
% \end{figure*}

\begin{figure}[t]
    \centering
        \includegraphics[width=\linewidth, trim=0 3 0 3, clip]{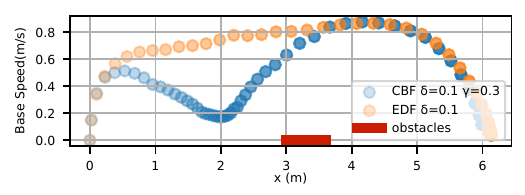}
        \vspace{-5mm} % Reduce vertical spacing if needed
        % \caption{Caption for figure 1}
        \label{fig:S-WBN-Speed}
    \caption{Comparing CBF against EDF on base speed during whole-body navigation around obstacles. Time progression is indicated by opacity. CBF promotes safer behaviours by significantly reducing speed near obstacles while maintaining nominal speed when far away. }
    \label{fig:S-WBN-TwoBoxes}
\end{figure}

\section{Experiments}\label{sec:experiments}
\revision{
We evaluated our proposed perceptive HTMPC framework with four sets of experiments both in simulation and on a real robot to test its safety and reactivity in semi-static unstructured environments.

Our robot consists of a Universal Robots UR10 arm mounted on a Clearpath Robotics Ridgeback omnidirectional base (\autoref{fig:main}). Two synchronized Orbbec Femto Bolt RGB-D cameras are installed, with one mounted on the base and the other on the end effector. An onboard laptop runs the entire perceptive HTMPC framework. Although equipped with two cameras, the FoV of our robot is highly limited compared to its size. Moreover, the hand camera's perception range is limited to 3~m to reduce artifacts due to wall reflections. 

We built our perception and mapping module on top of ORB-SLAM3~\cite{ORBSLAM3_TRO} and POCD~\cite{QianChatrathPOCD} and accommodated for the dual cameras setup. We also modified the state estimation method to use wheel odometry to improve robustness in featureless regions and extrapolate the estimated state to compensate for the low frame rate (15 FPS) and camera driver latency. State estimation ($50$ Hz) and dense mapping ($5$-$10$ Hz) run in two parallel threads to reduce the overall latency of the closed-loop system. 

The base trajectory is generated using a cutomized $A^*$ algorithm that finds the shortest path through multiple waypoints cascaded by a smoothening step. EE trajectory is a constant value at the desired goal waypoint for manipulation. 
The HTMPC controller is implemented using ACADOS~\cite{verschueren_acadosmodularopensource_2022}. The commanded joint velocity to the robot is calculated by interpolating the MPC solution at each control time step. 
The controller was set to run at $10$ Hz, and the commands are sent at $100$ Hz. 
The closed-loop delay of our system is measured to be around 0.5~s on average, including 0.13~s camera driver latency, 0.35~s in localization and mapping, and  0.03~s in control. 

To highlight the importance of object change detection in semi-static environments, we compared our proposed object-aware mapping against a typical approach that incorporate observations only at voxel level while assuming static enviroments (\autoref{sec:Exp-MS-WBN}). We also compared our proposed CBF safety constraints with the typical EDF formulation and quantified their performance on whole-body navigation under perception limitations using onboard cameras (\autoref{sec:Exp-S-WBN}). Lastly, we demonstrated reactivity of the close-looped system in safety-critical scenarios (\autoref{sec:Exp-R-WBN}) and showed its overall performance in sequential mobile manipulation applications (\autoref{sec:Exp-SequentialTasks}). A video of the experiments is available at the following link: \href{http://tiny.cc/peception-htmpc}{http://tiny.cc/peception-htmpc}.

\begin{figure}[t]
  \centering
  \begin{subfigure}[b]{\linewidth}
  \includegraphics[width=\linewidth, trim=0 7 0 0, clip]{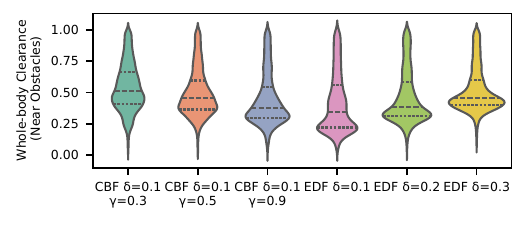}
  \caption{Whole-body clearance computed using ground truth map and exact robot geometry. Given the same safety margin, CBF exhibits larger traverse clearance than EDF.}\label{fig:S-WBN-Simulation-Clearance}
  \end{subfigure}
  \hfill
  \begin{subfigure}[b]{\linewidth}
\includegraphics[width=\linewidth, trim=0 7 0 0, clip]{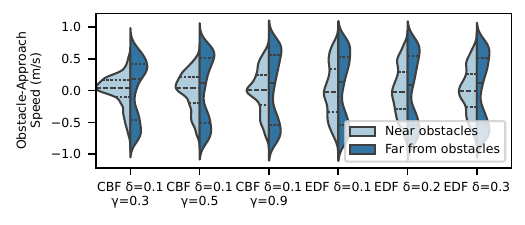}
  \caption{Obstacle-approach speed---base velocity projected onto the negative gradient of the ground truth map. CBF consistently limits the robot’s speed near obstacles, while EDF allows higher speeds even in close proximity, regardless of the safety margin settings.
}\label{fig:S-WBN-Simulation-Speed}
  \end{subfigure}
  % \begin{subfigure}[b]{\linewidth}
  %       % \includegraphics[width=\linewidth, trim=0 7 0 0, clip]{figure/benchmark/safety_margin.pdf}
  %       % \caption{Safety margin violation rates evaluated on both the perceived and ground truth maps, expressed as the percentage of control time steps. CBF exhibits consistent performance across both maps, while EDF shows a significantly higher violation rate on the ground truth map.}\label{fig:S-WBN-Simulation-SafetyMargin}
  %       \includegraphics[width=\linewidth, trim=0 2 0 0, clip]{figure/benchmark/collision_reasons.pdf}
  %       \caption{Safety margin violation rates evaluated on both the perceived and ground truth maps, expressed as the percentage of control time steps. CBF exhibits consistent performance across both maps, while EDF shows a significantly higher violation rate on the ground truth map.}\label{fig:S-WBN-Simulation-SafetyMargin}
  % \end{subfigure}
  \caption{Comparison of CBF and EDF within the HTMPC framework in the 45 simulated whole-body navigation trials under varying safety margin settings $\delta$ and $\gamma$ values.}\label{fig:S-WBN-Simulation}
\end{figure}
  \vspace{-2mm}

\subsection{Object-Aware vs Voxel-level Mapping}\label{sec:Exp-MS-WBN}
In this experiment, the robot is tasked to perform whole-body navigation in semi-static, unstructured environments where scene changes occur behind the robot, typically seen in long-horizon applications. As shown in \autoref{fig:S-WBN-Setup}, the robot needs to make a round trip between two waypoints. Immediately after it finishes the outbound trip (Phase 1), the scene was changed outside the robot's FoV. As a result, when returning (Phase 2), the robot needs to continuously detect changes in the environment to maintain a consistent map and update its motion plans in real time. 

We evaluated the proposed perceptive HTMPC framework with explicit change detection processes using object-level information against a baseline method that only performs voxel-level updates with static environments assumptions \cite{Rosinol20icra-Kimera}. Results show that the explict object-aware change detection benefits the system with a more accurate map, enabling more efficient task execution. An example is shown in \autoref{fig:MS-WBN} and three objects that were relocated are highlighted. Without explicit object-level change detection and with limited new observations, the baseline approach fails to clear out all previously occupied voxels by the relocated objects. Consequently, due to the persisting phantom obstacles, the robot must take a longer path. In contrast, as seen in \autoref{fig:MS-WBN-OBJ-CONF}, our proposed framework actively updates each object's consistency score as new observations come in, removes the objects when their consistency score $\mathbb{E}[v]$ drops below a set threshold. As a result, as shown in \autoref{fig:MS-WBN}, the proposed method keeps an artifact-free map and results in a shorter path found, enabling more efficient task execution.

\subsection{CBF vs EDF Safety Constraints}\label{sec:Exp-S-WBN}
We compare the proposed CBF-based safety constraint with the typical EDF-based formulation~\cite{pankert_perceptive_2020} within the HTMPC framework on whole-body navigation tasks.
First, we demonstrate the key differences between CBF and EDF safety constraints using a simple simulation example. The robot is tasked to navigate along a straight path while avoiding obstacles. As shown in \autoref{fig:S-WBN-Simulation-Speed}, unlike EDF, CBF significantly reduces speed near obstacles while maintaining nominal speed when far away, which improves task execution safety without significantly compromising efficiency. Moreover, CBF promotes higher traverse clearance than EDF given the same safety margin. In the following, we further demonstrate that these two behaviourial characteristics benefit the system with improved whole-body navigation performance when subjected to partial, delayed and inaccurate perceptions commonly seen in real-world applications.

%an essential safety criterion in human-centric environments where high speed near obstacles is perceived as unsafe---
\begin{figure}[t]
    \centering
    \begin{subfigure}[b]{\linewidth}
        \includegraphics[width=\linewidth, trim=0 3 0 8, clip]{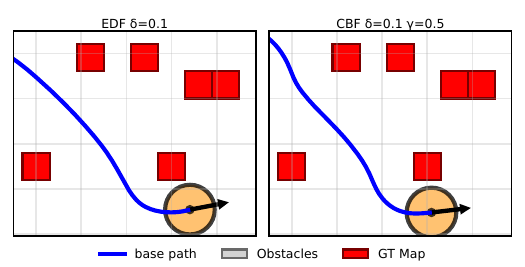}
        \vspace{-3mm} % Reduce vertical spacing if needed
        % \caption{Caption for figure 2}
        \label{fig:S-WBN-GT}
    \end{subfigure}\\
    \begin{subfigure}[b]{\linewidth}
        \includegraphics[width=\linewidth, trim=0 3 0 14.5, clip]{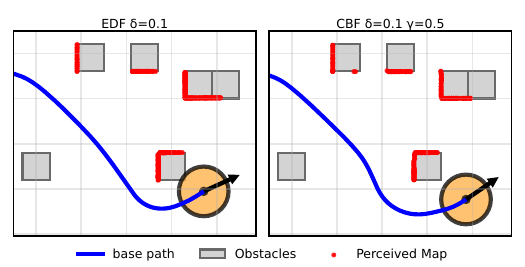}
        \caption{Qualitative navigation behaviours of EDF and CBF under GT and perceived maps.}
        \label{fig:S-WBN-Path}
        % \vspace{-1mm} % Reduce vertical spacing if needed
    \end{subfigure}\\
    \begin{subfigure}[b]{\linewidth}
        \includegraphics[width=\linewidth, trim=0 8 0 3, clip]{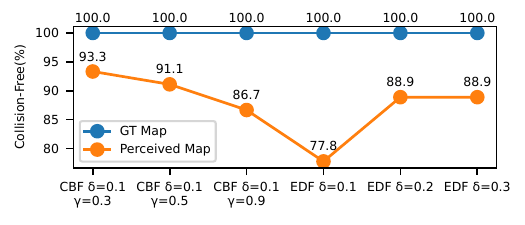}
        \caption{Average collision-free rate over the 45 simulated trials.}
        \label{fig:S-WBN-Collision-Rate}
    \end{subfigure}
    % \vspace{-10mm} % Reduce vertical spacing if needed
    \caption{Comparison of CBF and the EDF safety constraints on whole-body navigation under ground truth (GT) and perceived maps. Both methods are sufficient in collision avoidance when given complete maps. However, under incomplete maps, CBF is safer than EDF with the same safety margin parameter due to greater traverse clearance. }
    \label{fig:S-WBN-Clearance-on-Partial-Map}
    % \vspace{-10mm} % Reduce vertical spacing if needed
    
\end{figure}
\begin{figure}[t]
  \centering
    \includegraphics[width=0.9\linewidth, trim=5 2 5 6, clip]{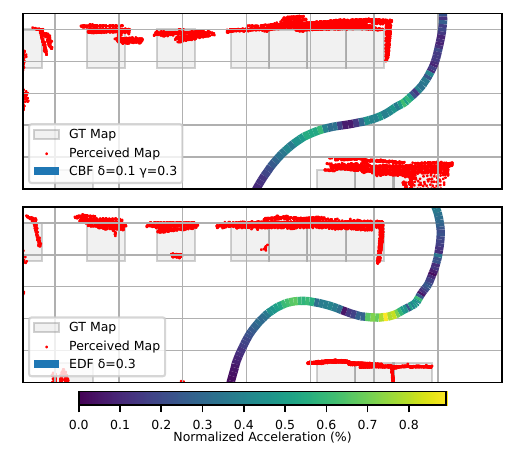}
  \caption{Real-world experiment comparing CBF and EDF within HTMPC, showing base paths observed in the experiments colored by base accelerations. Excessive safety margins in EDF $\delta=0.3$ trigger unnecessary constraint activations, increasing acceleration in tight spaces.}\label{fig:S-WBN-Realworld}
  \vspace{-2mm}
\end{figure}

\begin{figure}[t!]
      \centering
      \includegraphics[width=0.94\linewidth,trim={0 0cm 0 0}, clip]{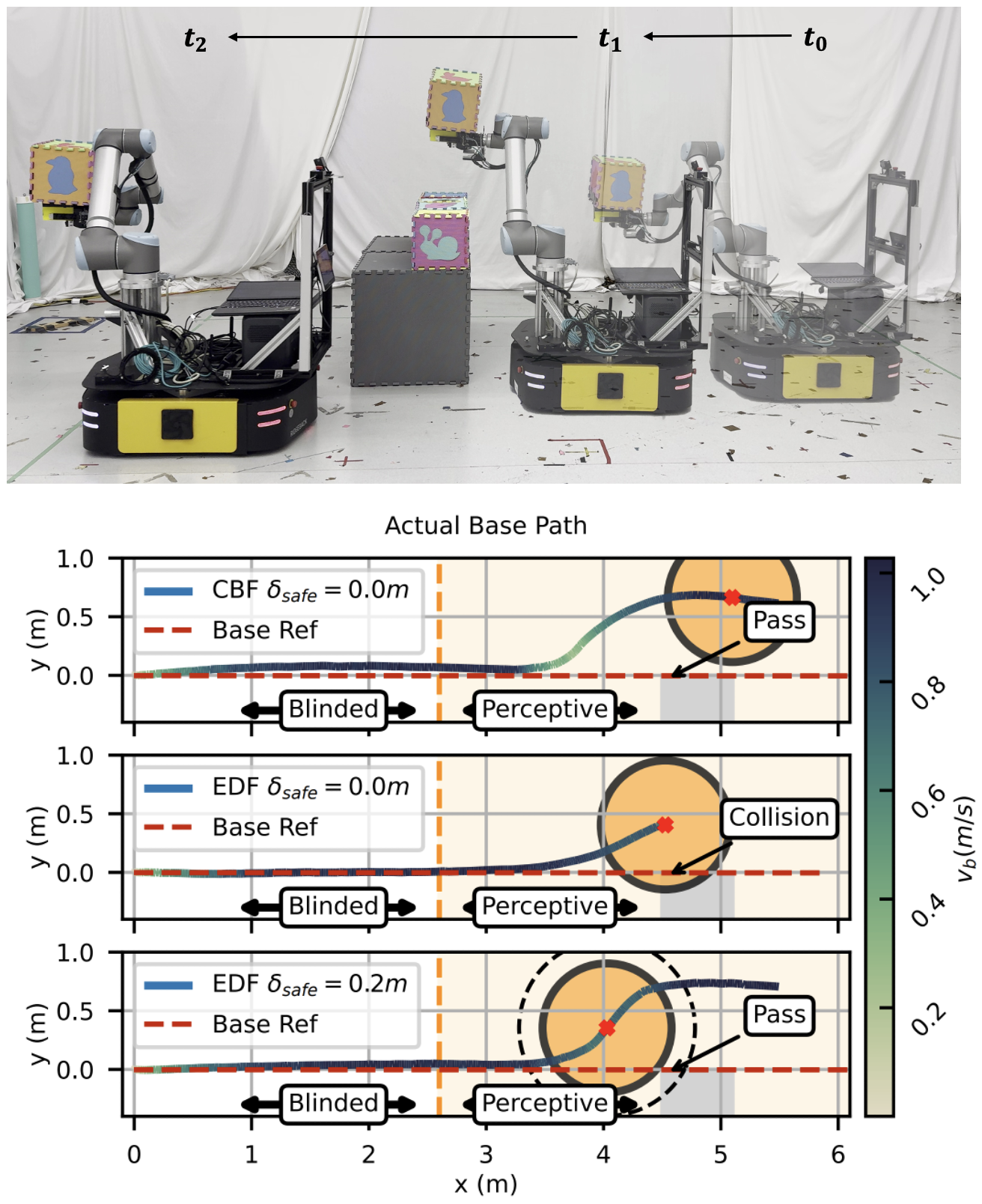}
      \caption{Real-world experiment comparing CBF and EDF safety constraints on whole-body navigation with challenging semi-static changes enforced by limited perception range. Time-lapse figure (top) and corresponding base paths (bottom) of the robot performing whole-body navigation around a box (shaded in grey). The robot was set to have a limited perception range of $1.9$ m and could only perceive the box when close to the box (shaded in yellow). CBF formulation enables the robot to safely drive passed and over the boxes, whereas the EDF formulation without an inflated safety margin results in a collision.}
      \label{fig:semi-static-exp-setup}
\end{figure}

First, we used the same whole-body navigation task described in \autoref{sec:Exp-MS-WBN}. The test includes one hand-designed real-world scenario as well as 15 randomly simulated scenarios, each with at least eight 0.6m boxes randomly placed in a 5m $\times$ 5m area, stacked in one or two layers (\autoref{fig:S-WBN-Setup}). For each simulated scenario, we test the system at 3 different desired speeds, or equivalently, reaction time, up to the robot's physical limit. The test totals 45 test scenarios. These scenarios reflect real-world perceptive control challenges such as incomplete maps due to limited FoV (\autoref{fig:S-WBN-Path}), inaccurate maps due to sensor noises (\autoref{fig:S-WBN-Realworld}), and short reaction time where conservative navigation behaviours near obstacles are preferred for safety. 

We performed ablation studies and compared the two approaches with different parameter settings and with the ground truth (GT) map and the perceived map for collision avoidance. We used several safety metrics for evaluation: traverse clearance, speed and acceleration near obstacles as well as collision-free rate.
The main observations are summarized below.

\begin{figure*}[t]
    \centering
    \begin{subfigure}[b]{\textwidth}
        \includegraphics[width=\linewidth, trim=0 8 0 8, clip]{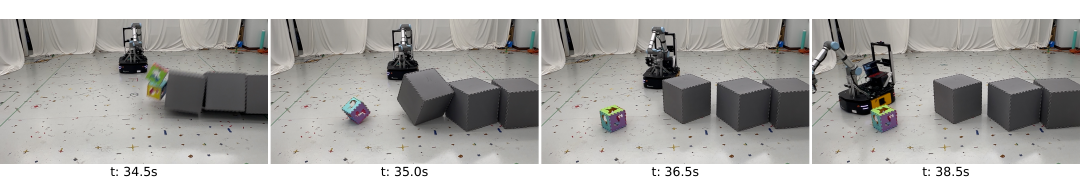}
        % \caption{Caption for figure 1}
    \vspace{-7mm} % Reduce vertical spacing if needed
    \end{subfigure}
    \\
    \begin{subfigure}[b]{\textwidth}
        \includegraphics[width=\linewidth, trim=0 3 0 8, clip]{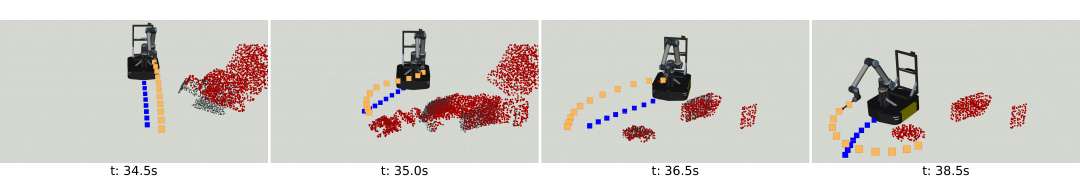}
        % \caption{Caption for figure 2}
    \end{subfigure}
    \caption{The reactive whole-body navigation task, highlighting real-time decision making with the proposed perceptive HTMPC approach in closed-loop with fast perception feedback. \textit{Top row}: time-series images from the experiment. \textit{Bottom row}: corresponding illustrations of raw point clouds (black), perceived map (red), and the HTMPC prediction EE (yellow) and base (blue) paths.}
    \label{fig:R-WBN}
    \vspace{-5mm}
\end{figure*}

Under identical safety margin settings, the proposed CBF framework consistently outperforms the EDF baseline in safety-related metrics. As shown in \autoref{fig:S-WBN-Simulation}, when $\delta = 0.1$, CBF better regulates speed and promotes larger traverse clearance when near obstacles while maintaining nominal speed when far. CBF's conservative behaviours reduces collision risks under partial observations as shown in \autoref{fig:S-WBN-Path}. More specifically, both approaches are effective for collision avoidance when a ground-truth (GT) map is available. However, when the map is incomplete, CBF benefits from its larger clearance with reduced exposure to both observed and unobserved part of the obstacles. In contrast, EDF operates with minimal clearance, which increases the risk of collisions. This trend holds across all 45 simulated trials: \autoref{fig:S-WBN-Collision-Rate} shows that CBF yields a higher percentage of collision-free cases than EDF with the same safety margin.

For CBF, the level of reduced traverse speed and increased clearance can be tuned via the parameter $\gamma$. As $\gamma$ increases, CBF's safety performance degrades toward EDF levels (\autoref{fig:S-WBN-Simulation}, \autoref{fig:S-WBN-Collision-Rate}), providing a tunable trade-off between safety and efficiency. While EDF can improve collision avoidance performance by increasing the safety margin $\delta$ (\autoref{fig:S-WBN-Collision-Rate}), it still permits unsafe speeds near obstacles (\autoref{fig:S-WBN-Simulation-Speed}). 
Moreover, excessive safety margins in EDF may trigger unnecessary constraint activations, increasing acceleration in tight spaces and decreasing the robot's traversability (\autoref{fig:S-WBN-Realworld}).

We further verified our conclusions by testing the robot with sudden detection of new objects when subjected not only to a limited FoV but also the perception and actuation delays of the real robot. The experiment setup and an example robot behaviour are shown in \autoref{fig:semi-static-exp-setup}. The robot base was commanded to drive along a straight line with a fixed EE waypoint in the base frame to its right. Boxes were placed on its base and arm paths, so the robot needed to move both the arm and base to drive past the obstacles. 
To simulate the sudden detection of new obstacles in semi-static environments, we severely decreased the robot's perception range in our experiments. Qualitative results are shown in \autoref{fig:semi-static-exp-setup}.

% We compared CBF and EDF with different parameters on 6 trials. Qualitative results are shown in \autoref{fig:semi-static-exp-setup} and statistical results are tabulated in \autoref{tab:semi-static-collision-free}. 

Same as our previous conclusions, when the safety margin is identical, CBF approach enables the robot to pass the obstacle safely whereas the EDF approach results in collision. By directly limiting the robot's approach speed towards the obstacles as well as encouraging larger traverse clearance, CBF becomes less susceptible to perception and control delays and inaccuracy. Although increasing safety margin for EDF can increase traverse clearance, it will necessarily compromise traversability as seen in \autoref{fig:S-WBN-Realworld}. In the following section, we will further demonstrate the reactivity of the proposed perceptive HTMPC framework in challenging real-world scenarios.

\subsection{Perceptive-HTMPC for Reactive Whole-Body Navigation}\label{sec:Exp-R-WBN}
In this real-world experiment, we demonstrate the reactivity of the proposed perceptive HTMPC framework with CBF-based safety constraints in safety-critical scenarios where dynamic obstacles appear suddenly in front of the robot, requiring rapid responses to unforeseen hazards. As shown in \autoref{fig:R-WBN}, the robot is tasked with reaching a waypoint at its maximum speed of 1m/s. During execution, stacks of boxes are suddenly pushed and scattered in front of the robot. Due to the high speed and short distance to collision (DTC), the system must perform mapping, localization, and collision avoidance online, within a very short time window. 

To assess performance under different levels of environmental complexity, we conducted experiments with two clutterness levels: two-box and five-box scenarios, running 15 trials for each. In the two-box case, the HTMPC controller followed a fixed path for both the EE and the base. In contrast, the five-box case employed reactive local planners due to the increased complexity in the environment. To evaluate the robot’s reactive behaviour, we varied the DTC by introducing the disturbances at different time instances.

The minimum DTC at which successful avoidance was observed was 0.6~m for the EE and 1.48~m for the base. Considering computation delays in the closed-loop system, the theoretical minimum DTC is approximately 0.5~m at a speed of 1~m/s, excluding the low-level joint velocity tracking delays. The proposed perceptive HTMPC is able to resolve collisions with dynamic obstacles with a minimum DTC close to this theoretical bound. 
Overall, the robot achieved a collision-free rate of $87\%$ in the two-box test and $60\%$ in the more challenging five-box test, where the boxes rolled faster and further, and approached the robot more closely.
All observed collisions were attributed to delays and incompleteness in the perceived map where collision avoidance became physically impossible.
}

\subsection{Perceptive-HTMPC for Long-Term Sequential Tasks}\label{sec:Exp-SequentialTasks}
In this experiment, we demonstrate the proposed perceptive HTMPC framework for a long-term sequential mobile manipulation in semi-static environments. As shown in \autoref{fig:main}, the robot is tasked to navigate along a circular aisle (red) three times while receiving objects at two locations (green). When the robot is running, boxes (orange) are rearranged to test the robot's ability to detect changes, update the map, and react via re-planning and reactive control. 

The left and right panels of~\autoref{fig:main} highlight the robot's response to the changed boxes between visits. Upon observing and associating the changed boxes to their previous reconstructions, the perception module updates the state model of each box. Once sufficient measurements are accumulated, the boxes are determined to be changed and hence reconstructed at their newly observed locations. In reaction to the changed boxes, the robot had to deviate from its reference base, retract or lift up its arm for a safe passage.

This experiment also showcase how HTMPC leverages the robot's kinematic redundancy in sequential mobile manipulation tasks. For whole-body navigation, the HTMPC has a cost function $[\mathcal{J}_{\mathit{base}}, \mathcal{J}_{\mathit{EE}}]$, where the primary base tracking task is complemented with a secondary EE task, which allows the robot's hand camera to look at a future point on the base path to gain earlier perception information for collision avoidance. For manipulation, the HTMPC has the same set of cost functions but in a reversed order, $[\mathcal{J}_{\mathit{EE}}, \mathcal{J}_{\mathit{base}}]$, enabling the robot to perform manipulation tasks on a moving base for improved efficiency.

\section{Conclusions and Future Works}\label{sec:conclusion}
We presented a perceptive HTMPC framework for sequential mobile manipulation in unstructured semi-static environments. 
Extensive simulations and real-robot experiments demonstrate that our proposed perceptive HTMPC framework enables accurate mapping and reliably adapts to semi-static changes in the environments during long-term operation while exploiting the robot's redundancy for sequential tasks execution.

% We identified several improvements that could be made in future works. 
We see several promising directions for future work. First, the robot's safety  depends on map accuracy; in this work, we designed a strategy that directs the hand camera towards a future point on the planned navigation path to improve camera coverages. Future work could leverage active perception based methods to strategically place hand camera to minimize map uncertainties and therefore improve the robot's safety. Another possible future direction is integrating language-based planners to solve more complex tasks than pick-and-place and/or facilitate intuitive human-robot interaction, ultimately demonstrating these capabilities in real-world human-centric applications.

\bibliographystyle{IEEEtran}
		\bibliography{sections/references}

\addtolength{\textheight}{-12cm}   % This command serves to balance the column lengths

% \input{sections/biography}
\end{document}